\documentclass[10pt,twocolumn,letterpaper]{article}

\usepackage{cvpr}              
\definecolor{cvprblue}{rgb}{0.21,0.49,0.74}
\usepackage[pagebackref,breaklinks,colorlinks,allcolors=cvprblue]{hyperref}

\usepackage[utf8]{inputenc} 
\usepackage[T1]{fontenc}    
\usepackage{hyperref}       
\usepackage{url}            
\usepackage{booktabs}       
\usepackage{amsfonts}       
\usepackage{nicefrac}       
\usepackage{microtype}      
\usepackage{xcolor}         
\usepackage{colortbl}
\definecolor{customcolor}{HTML}{D3D3D3}
\usepackage{amsmath} 
\usepackage{multirow}
\usepackage{microtype}
\usepackage{graphicx}

\usepackage{booktabs} 
\usepackage[ruled,vlined,linesnumbered]{algorithm2e}
\usepackage{colortbl}
\usepackage{wrapfig}

\def\paperID{} 
\def\confName{CVPR}
\def\confYear{2026}

\title{Geometric-Aware Hypergraph Reasoning 

for Novel Class Discovery in Point Cloud Segmentation}

\author{Zihao Zhang\textsuperscript{1}, Aming Wu\textsuperscript{2}\thanks{Corresponding author.}, Yang Li\textsuperscript{1}, Yahong Han\textsuperscript{1}, Jialie Shen\textsuperscript{3}\\
\textsuperscript{1}School of Artificial Intelligence, College of Intelligence and Computing, Tianjin University, China\\
\textsuperscript{2}School of Computer Science and Information Engineering, Hefei University of Technology, China\\
\textsuperscript{3}Department of Computer Science
City St George’s, University of London, UK\\
{\tt\small zhangzihao2490@tju.edu.cn, liyang1389@tju.edu.cn, yahong@tju.edu.cn,} \\
{\tt\small amwu@hfut.edu.cn, jerry.shen@citystgeorges.ac.uk}
}

\begin{document}
\maketitle
\vspace{-10pt}
\begin{abstract}
Novel Class Discovery in Point Cloud Segmentation is recently proposed, aiming to leverage knowledge from known classes to automatically segment unlabeled classes within point clouds. 
The core of this task lies in leveraging the geometric and semantic knowledge of multiple known classes to achieve semantic understanding and segmentation of novel classes.
However, existing methods overlook the high-order associations between known and novel classes, relying solely on binary associations for class assignment and novel class reasoning, which leads to less precise semantic segmentation.
To address these issues, we introduce a hypergraph structure to model high-order associations among classes, enabling collaborative reasoning from known classes to novel classes, extending beyond traditional binary relations.
Additionally, existing methods focus excessively on extracting semantic information when processing point cloud data, neglecting the importance of geometric features. To address this, we introduce a novel prototype, Geometric-Aware Prototypes, enhancing the model's ability to capture geometric spatial information.
By propagating geometric information through hyperedges, our method enhances the understanding of spatial distributions across classes, improving segmentation accuracy.
Significant performance improvements achieved on the SemanticKITTI and SemanticPOSS datasets demonstrate the superiority of our method. Our code is available at \url{https://github.com/2490o/HyperNCD}.
\end{abstract}    
\vspace{-10pt}
\section{Introduction}

\label{sec:intro}

\begin{figure}[!t]
	\centering
        
	\includegraphics[width=\columnwidth]{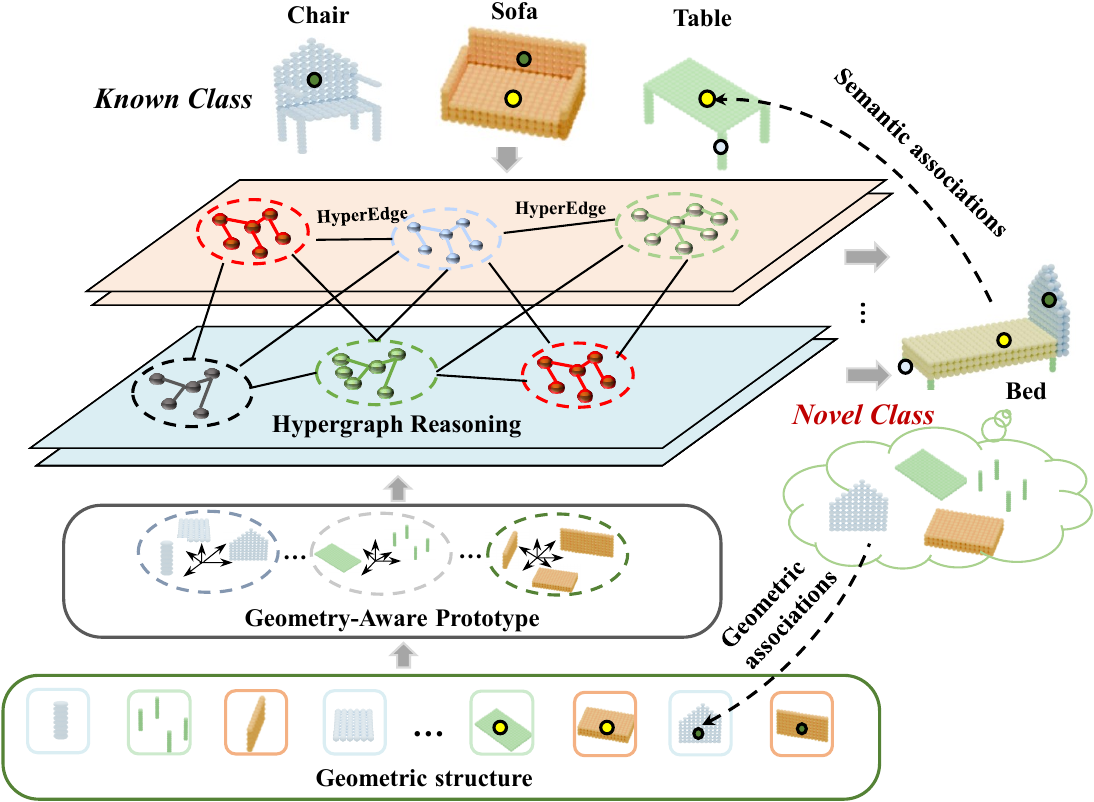}
	\DeclareGraphicsExtensions.
	\begin{center}
		\vspace{-5pt}
		\caption{The core of Geometric-Aware Hypergraph Reasoning lies in constructing a hypergraph structure to facilitate collaborative reasoning for novel classes. For the novel class ‘bed,’ hyperedges connect known class prototypes such as ‘chair,’ ‘sofa,’ and ‘table.’ Through cross-category information associations, the model can reason the novel class based on known geometric and semantic information, even without direct training on the ‘bed’ class.
        }\label{fig-1}
        \vspace{-15pt}
	\end{center}
	\vspace{-15pt}
\end{figure}

With the rapid development of 3D perception technology \cite{3D1,point1}, point cloud semantic segmentation has shown significant value in fields such as autonomous driving \cite{vggdrive, autonomousdriving2, autonomousdriving3} and intelligent robotics \cite{driving, driving2}. However, most existing methods assume that all semantic classes are known during training, which poses clear limitations in open-world scenarios. To address this, Riz et al. \cite{NOPS} introduce the Novel Class Discovery (NCD) framework into point cloud semantic segmentation, enabling the segmentation of novel classes in point clouds by leveraging knowledge from known classes. This makes NCD for 3D semantic segmentation more valuable for research and applications compared to traditional Point Cloud segmentation methods.

For the task of NCD in 3D semantic segmentation, the core challenge lies in leveraging higher-order associations from multiple known classes to reason novel classes, including geometric and semantic associations. The NOPS \cite{NOPS} pioneered an approach based on online clustering and uncertainty estimation. However, this method relies on binary relations between classes for information associations and label assignment, which limits its ability to capture the high-order associations among multiple classes.
DASL \cite{DLAS} introduces regional consistency and a semi-relaxed optimal transport algorithm to address class imbalance in point cloud data. However, its reliance on semantic associations between points and regions restricts information association to pairs of closely related classes or regions, limiting the associations of information across multiple classes.

To address these issues, we introduce the hypergraph structure to model high-order associations among classes, which facilitates collaborative reasoning for novel classes. Unlike traditional graph structures, hypergraphs allow the connection of multiple nodes within a hyperedge, effectively capturing higher-order associations among classes rather than being restricted to binary relations. This method enhances the associations of information across multiple classes, supporting collaborative reasoning for novel class discovery. By leveraging high-order associations between novel classes and multiple konwn classes, the model can better understand the semantic and geometric context of novel categories.
Furthermore, we propose a novel prototype, the Geometric-Aware Prototype, which captures the geometric and spatial characteristics of different classes in point cloud data. This addresses the limitations of existing methods, which emphasize semantic features while overlooking the essential 3D geometric features of point clouds.

Specifically, as illustrated in Figure \ref{fig-1}, we propose a method, Geometric-Aware Hypergraph Reasoning for Novel Class Discovery in Point Cloud Segmentation. Our method surpasses previous methods that rely on pairwise class or region information associations for label assignment and information transmission. Instead, we employ a hypergraph structure to model high-order associations between various classes, allowing for richer information associations and interaction that supports collaborative reasoning for novel classes.
Our method begins by extracting both local and global spatial information from point cloud features to construct geometric-aware prototypes, which are then represented as nodes in the hypergraph, as shown in the lower part of Figure \ref{fig-1}. Using hyperedges constructed based on geometric and semantic similarity, we model the high-order associations between different prototypes, capturing latent geometric connections and semantic associations among different classes. By connecting related prototypes through hyperedges, the model gains access to richer contextual semantic and geometric information during learning. This high-order information association not only enables collaborative reasoning of novel classes using multiple known class prototypes but also enhances understanding of known classes, thereby reducing ambiguity in classification, as shown in the middle part of Figure \ref{fig-1}.

Finally, we design a dynamic hyperedge construction mechanism that adjusts the hyperedge structure based on inter-prototype associations across training batches. This mechanism ensures that the model can continually optimize high-order associations between prototypes, thereby mitigating class imbalance issues across different batches.

To demonstrate the effectiveness of our method, we conducted extensive experiments on two widely used datasets: SemanticKITTI and SemanticPOSS. The experimental results validate the effectiveness of our proposed method.





\section{Related Works}

\label{sec:formatting}

\subsection{Novel Class Discovery}

While the NCD task \cite{NCD2, JLCR} has seen progress in the 2D image domain \cite{2D1, 2D2, 2D3, 2D4}, research on 3D point cloud NCD \cite{NOPS} remains limited. NCD leverages known classes to infer semantics for unknown classes and follows two main approaches. The first category \cite{NCD1} trains a classification network on known data, then clusters novel data based on predictions or network features, but often overfits base classes, limiting performance on novel classes. The second category \cite{NCD4, NCD5} includes both known and novel samples in clustering to promote shared feature representations, enhancing generalization. However, it relies on binary relationships and clusters by feature similarity, missing high-order class relationships. To address this, we introduce a hypergraph \cite{hy1, hy2} structure to capture high-order interactions between classes. This structure enables cooperative reasoning about novel classes by linking them with multiple base classes, enriching semantic and geometric understanding.

 \begin{figure*}[!t]
  \centering
  \includegraphics[width=2.0\columnwidth]{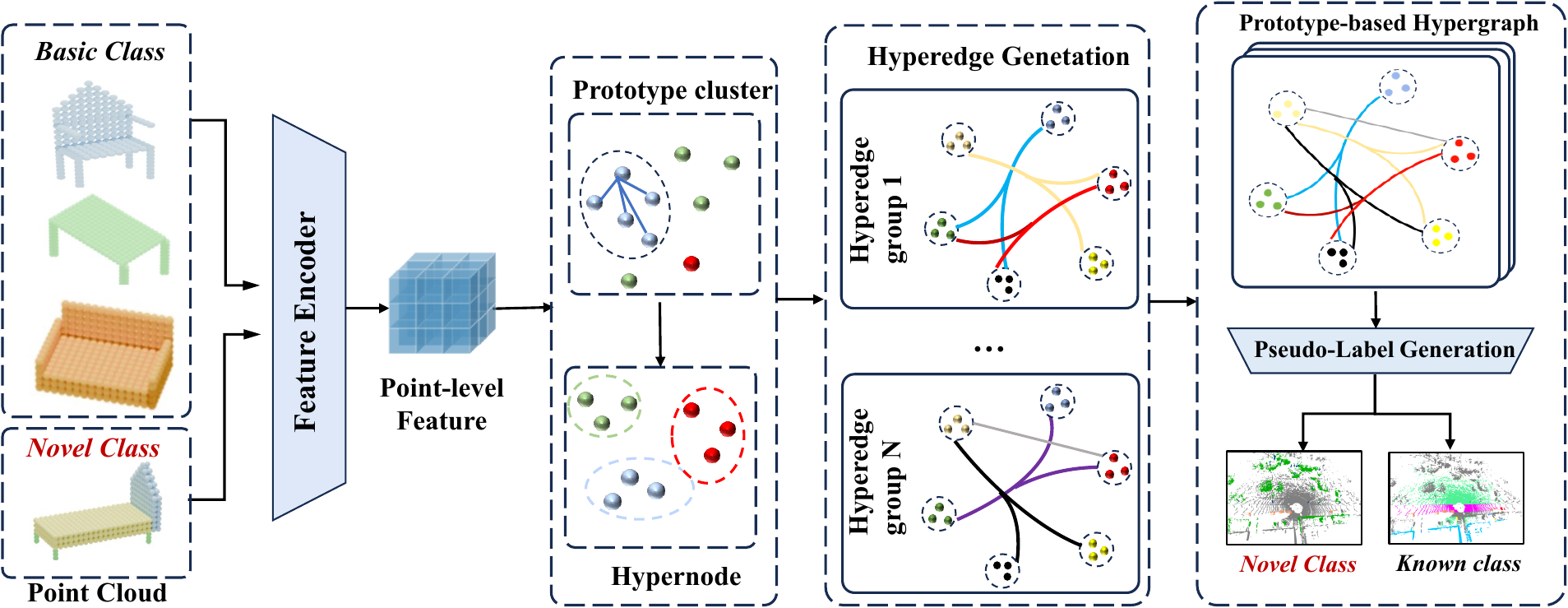}
  \vspace{-5pt}
  \caption{\textbf{The overall architecture}. The point cloud data is first fed into a backbone network to extract point-wise features $F$, which are then clustered to generate geometric-aware prototypes $P_i$. These prototypes serve as nodes in a hypergraph, and hyperedges $e_i$ are constructed based on the similarity of their geometric and semantic features, forming a hypergraph structure within each batch. A dynamic hyperedge adjustment mechanism is subsequently introduced to update the hyperedges in real time according to the evolving feature distribution across multiple batches, while simultaneously generating pseudo-labels for emerging novel classes.}
    \label{F2}
     \vspace{-10pt}
\end{figure*}

\subsection{Point Cloud Segmentation}

With advancements in 3D perception, point cloud semantic segmentation \cite{Point2, point} has shown substantial value in areas like autonomous driving \cite{autonomousdriving3, autonomousdriving4, autonomousdriving2} and robotics \cite{driving, driving2}. However, most methods \cite{3d11,3d22} assume all semantic categories are known during training, which limits applicability in open-world scenarios \cite{OVD, SECOT, S_DG}. Open-world perception requires recognizing known classes and learning novel, unlabeled ones. Riz et al. \cite{NOPS} introduced the Novel Class Discovery (NCD) task to point cloud segmentation, using online clustering and uncertainty estimation. Yet, traditional clustering methods often miss high-order inter-class interactions, especially with intricate novel-class relationships and imbalanced data. DASL \cite{DLAS} partially addressed these issues with regional consistency and a semi-relaxed optimal transport algorithm, but it relies mainly on point-to-region relationships, which limit its ability to capture high-order class interactions. Both methods neglect 3D spatial and geometric properties, leading to suboptimal segmentation. 

\section{Method}
The overall framework of our method, Geometric-Aware Hypergraph Reasoning, is shown in Figure \ref{F2}. The core idea is to leverage Geometric-Aware Prototypes to construct a hypergraph structure that captures the high-order interactions between multiple classes, aimed at effective novel class discovery. Additional details on the problem definition and data settings are available in the Supplementary Material.

\subsection{Geometry-Aware Prototype}
We propose a Geometry-Aware Prototype that captures both spatial and structural features of different categories in point cloud data. This prototype generation is done through a module that maps point cloud features to multiple prototypes, incorporating geometric awareness.
Given an input point cloud, we utilize a feature extraction network to obtain a feature map \( F \in \mathbb{R}^{N \times P \times C} \), where \( N \) represents the batch size,  \( P \) is the number of points, and \( C \) denotes the number of channels. Due to the sparse and unordered nature of point cloud data, we first reshape $F$ to $F_{flat} \in \mathbb{R}^{N \cdot P \times C}$ and apply a 1D convolution to map the feature map into $K$ clusters, corresponding to the number of prototypes:
\begin{eqnarray}
\sigma = \text{Conv1d}(F_{flat}),
\end{eqnarray}
where $\sigma \in \mathbb{R}^{N \cdot P \times K}$ represents the output of this convolution.
Next, we apply the Softmax function along the $K$ dimension of $\sigma$ to obtain the soft assignment matrix $S_p$:
\begin{eqnarray}
S_p = \text{softmax}(\sigma),
\end{eqnarray}
where $S_p \in \mathbb{R}^{N \cdot P \times K}$ represents the weight of each point in the batch assigned to each prototype. We denote the element in the $i$-th row and $k$-th column of $S_p$ as $s_{i,k}$, which is the soft assignment weight of the $i$-th point to the $k$-th prototype.
Each prototype $P_k \in \mathbb{R}^{C}$ is designed to simultaneously capture semantic and geometric information. During the feature extraction stage, the feature $F_i$ for the $i$-th point $x^i$ in the batch is constructed by concatenation:
\begin{eqnarray}
F_i = [f_{\text{sem}}(x^i); f_{\text{geo}}(x^i)],
\end{eqnarray}
where $f_{\text{sem}}(x^i)$ and $f_{\text{geo}}(x^i)$ represent the semantic and geometric features, respectively. The dimension $C$ of the prototype $P_k$ is equal to the sum of the dimensions of these two feature components.
To extract the geometric feature $f_{\text{geo}}(x^i)$, we first construct the local neighborhood $\mathcal{N}(x^i)$ using the K-Nearest Neighbors (KNN) algorithm with $K=15$. This feature is then computed based on the eigenvalues of the Euclidean distance covariance matrix of $x^i$ and its neighbors, specifically by extracting structural cues like Linearity, Planarity, and Scattering, which quantify local curvature. Further details on the quantification of local curvature are elaborated in the supplementary material. 
In parallel, the semantic feature $f_{\text{sem}}(x^i)$ is derived by applying multiple layers of 3D convolution over $\mathcal{N}(x^i)$, facilitating the progressive aggregation of high-level contextual information.

To update the $k$-th prototype $P_k$, we first compute the soft-weighted residual for each point $x^i$ with respect to $P_k$. We flatten the input feature $F \in \mathbb{R}^{N \times P \times C}$ into a matrix of total dimension $(N \cdot P) \times C$, and broadcast $P_k$ to match the dimension of the flattened features. For the prototype $P_k$, its weighted residual sum $R_k$ is defined as:
\begin{eqnarray}
R_{k} = \sum_{i=1}^{N \cdot P} s_{i,k} \cdot \big(F_{\text{r}}^i - P_k\big),
\end{eqnarray}
where $F_{\text{r}}^i \in \mathbb{R}^{C}$ is the feature vector of the $i$-th point from the flattened feature matrix, and $s_{i,k}$ is the element from the soft assignment matrix $S_p$.
Subsequently, the updated prototype $P_k'$ is obtained by performing $L_2$-norm normalization on the residual sum:
\begin{eqnarray}
P_{k}' = \frac{R_{k}}{\left\Vert R_{k} \right\Vert_2},
\end{eqnarray}
where $\Vert \cdot \Vert_2$ denotes the Euclidean distance. This update mechanism enables the prototypes to dynamically converge towards the centers of the spatial and semantic patterns present in the point cloud data.

\begin{algorithm}[t]
   \caption{Geometry-Aware Hypergraph Construction and Novel Class Discovery}
   \KwIn{Training batch \( b \), Prototypes \( \{P_1, \dots, P_K\} \), Similarity weight \( \alpha \)}
   \KwOut{Updated hypergraph \( \mathcal{H} \), Pseudo-labels \( \hat{y}_i^{(b)} \) for novel prototypes}

   \textbf{Dual Similarity Computation:} \\
   \ForEach{prototype pair \( (P_i, P_j) \)}{
       Compute similarity: \(S_b(P_i, P_j) = \alpha S_g + (1{-}\alpha) S_s\)
   }

   \textbf{Initial Hyperedge Construction:} \\
   \ForEach{prototype \( P_i \)}{
       Select top \( M \) similar prototypes \(\{P_{j_1}, \dots, P_{j_M}\}\)\;
       Form hyperedge \( e_i = \{P_i, P_{j_1}, \dots, P_{j_M}\} \)
   }
   Set \(\mathcal{E}^{(b)} = \{e_i^{(b)} \,|\, i = 1, \dots, K\},\)

   \textbf{Dynamic Hyperedge Adjustment:} \\
   \ForEach{hyperedge \( e_i^{(b)} \)}{
 \( w(e_i^{(b)}) = \frac{1}{M}\sum_{P_j\in\mathcal{N}_i} S_b(P_i, P_j) \)
   }

   \textbf{Novel Class Discovery:} \\
   \ForEach{novel prototype \( P_i \)}{
       Compute similarity to known prototypes \(S_b(P_i, P_j)\)\;
       \uIf{similarity exceeds threshold}{
           Assign pseudo-label: \( \hat{y}_i^{(b)} = \arg\max_j S_b(P_i, P_j) \)
       }
       \Else{
           Create a novel class label and update the prototype:
           \( P_i^{(b+1)} = P_i^{(b)} + \gamma(x_u - P_i^{(b)}) \)
       }
   }

   \textbf{Hypergraph Update:} \\
   Recompute \( S_b \) and rebuild \( \mathcal{E}^{(b+1)} \)

   \Return{Updated hypergraph \( \mathcal{H} = (\mathcal{V}, \mathcal{E}^{(b+1)}) \), pseudo-labels \( \hat{y}_i^{(b)} \)}
\end{algorithm}

\subsection{Hypergraph Structure Construction}

After obtaining the clustered geometry-aware prototypes, we calculate the distance between prototypes by combining both geometric and semantic feature similarities, and use this similarity measure to construct hyperedges, as shown in Algorithm 1.
In each training batch \( b \), to more effectively fuse geometric and semantic information, we compute prototype similarity by considering both local curvature and overall global structure:
\begin{eqnarray}S_b(P_i, P_j) = \alpha \cdot S_{\text{g}}(P_i, P_j) + \beta \cdot S_{\text{s}}(P_i, P_j),
\end{eqnarray}  
where \( S_{\text{g}} \) measures geometric similarity, \( S_{\text{s}} \) measures semantic similarity, \( \alpha \) is the weight coefficient that balances geometric and semantic information, and \( \beta = 1 - \alpha \). $P_i$ and $P_j$ represent the prototype vectors updated or utilized in the current batch.
To compute the geometric similarity $S_{\text{g}}$, we first define the Average Local Distance $D_{\text{g}}$ by capturing the distribution differences between the local neighborhoods associated with the prototypes:
\begin{eqnarray}
D_{\text{g}}(P_i, P_j) = \frac{1}{|N_i| \cdot |N_j|} \sum_{x^p \in N_i} \sum_{x^q \in N_j} \left\Vert x^p - x^q \right\Vert_2,
\end{eqnarray}  
where $N_i$ and $N_j$ denote the sets of original points associated with the prototypes $P_i$ and $P_j$ in batch $b$, respectively. The distance $D_{\text{g}}$ is computed as the average Euclidean distance between points in these neighborhoods. We then transform this distance into the normalized similarity score $S_{\text{g}}$ using a Gaussian kernel:
\begin{eqnarray}
S_{\text{g}}(P_i, P_j) = \exp \left( -\frac{D_{\text{g}}(P_i, P_j)}{\tau} \right),
\end{eqnarray} 
where $\tau$ is a temperature hyperparameter, which we set to 0.1 in our experiments. This metric effectively quantifies the spatial proximity and structural overlap between the regions represented by the two prototypes.

Semantic features measure the distribution differences of prototypes in the feature space. We quantify their semantic similarity using cosine similarity:
\begin{eqnarray}
S_s(P_i, P_j) = \frac{P_i \cdot P_j}{\left\Vert P_i \right\Vert_2 \left\Vert P_j \right\Vert_2},
\end{eqnarray}
where $P_i$ and $P_j$ are prototypes from different categories, $\cdot$ denotes the dot product, and $\Vert \cdot \Vert_2$ is the Euclidean distance.

Based on the similarity measure, connect multiple similar prototypes to form a hyperedge. Specifically, for each prototype $P_i$, select its $M$ nearest prototypes $\{P_{j_1}, P_{j_2}, \dots, P_{j_M}\}$ and construct a hyperedge $e_i$ containing these nodes:
\begin{eqnarray}e_i = \{P_i, P_{j_1}, P_{j_2}, \dots, P_{j_M}\}.\end{eqnarray}
To avoid overly complex connections, we introduce a hyperparameter $M$ to limit each prototype’s connected neighbors, preventing redundancy in hypergraph construction. Experiments show optimal performance at $M = 8$. This approach enables a hypergraph structure that effectively captures high-order inter-class associations, enhancing the model’s representation of complex interactions.

\subsection{Dynamic Hyperedge Adjustment Mechanism}
In an open-world scenario, a fixed hyperedge structure limits the recognition of novel classes. To address this issue, we propose a dynamic hyperedge adjustment mechanism that integrates geometric similarity and multi-class relationships for effective novel class discovery.
Based on the computed similarity measure, we first construct the hyperedges and obtain the initial hyperedge set \( \mathcal{E}^{(b)} \):
\begin{eqnarray}
\mathcal{E}^{(b)} = \{e_i^{(b)} \,|\, i = 1, \dots,K \},
\end{eqnarray}
where $\mathcal{E}^{(b)}$ represents the updated hyperedge set at batch $b$, and $K$ is the total number of prototypes. Each hyperedge $e_i^{(b)}$ contains the prototype $P_i$ and its $M$ most related prototypes, based on geometric and semantic similarities. This comprehensive set reflects the higher-order relationships among prototypes in the current batch.
Next, we introduce a dynamic weighting mechanism. The weight $w(e_i^{(b)})$ for each hyperedge is defined as follows:
\begin{eqnarray}
w(e_i)=\frac{1}{M}\sum_{P_j\in\mathcal{N}_i}\Big[\alpha\, S_g(P_i,P_j)+(1-\alpha)\, S_s(P_i,P_j)\Big],
\end{eqnarray}
where $\alpha$ is the same weight as in Eq.~(6) and $|\mathcal{N}_i|=M$. This dynamic scheme enhances the flexibility and expressiveness of the hypergraph, enabling more accurate modeling of complex inter-class dependencies.

\subsection{Hypergraph for Novel Class Discovery}

After constructing the hypergraph structure, we further leverage its high-order inter-class relationships to effectively discover novel 3D categories. Specifically, the constructed hypergraph is used to generate pseudo-labels. For a novel prototype $P_i$ encountered in batch $b$, we compute its similarity with all existing prototypes in the current batch using the previously defined similarity measure $S_b(P_i, P_j)$. If its similarity with a known prototype exceeds a predefined threshold, it is assigned the pseudo-label of that class. Otherwise, it is considered a novel category and assigned a new label.
The pseudo-label assignment can be formalized as:
\begin{eqnarray}
\hat{y}_i^{(b)} = \arg\max_{P_j \in \mathcal{V}} S_b(P_i, P_j),
\end{eqnarray}
where $\hat{y}_i^{(b)}$ denotes the pseudo-label assigned to prototype $P_i$, and $\mathcal{V}$ represents the current set of prototypes.
If the prototype is identified as belonging to a novel class, it is incorporated into the training set, and its geometric and semantic features are updated accordingly. The prototype update rule is given by:
\begin{eqnarray}
P_j^{(b+1)} = P_j^{(b)} + \gamma \cdot (x_u - P_j^{(b)}),
\end{eqnarray}
where $\gamma$ is the learning rate controlling the update step, $x_u$ is the feature vector of the novel sample, and $P_j^{(b+1)}$ denotes the updated prototype in the $(b+1)$-th training batch.

Subsequently, based on the updated prototype features, we recompute the similarity scores and dynamically adjust the hyperedge structure:
\begin{eqnarray}
\mathcal{E}^{(b+1)} = \{e_i^{(b+1)} \,|\, i = 1, \dots, K \},
\end{eqnarray}
where $e_i^{(b+1)}$ denotes the hyperedge reconstructed for prototype $P_i$ in the $(b+1)$-th batch, based on the updated similarity measures.
This dynamic mechanism allows the hypergraph to adapt to the emergence of novel classes, thereby enhancing the model’s ability to capture evolving category structures and improving its novel class discovery performance in open-world settings. 
To enhance the robustness of pseudo-labels, we incorporate a label smoothing strategy. Detailed implementation and the corresponding loss function are provided in the Supplementary Material.

\begin{table*}  
  \centering
  \vspace{-15pt}
  \caption{The novel class discovery results on the SemanticPOSS dataset. ‘Spilt’ denotes the number of splits. ‘Full’ denotes the results obtained by supervised learning. The gray values are the novel classes in each split.}
  \fontsize{22}{22}\selectfont
     \resizebox{1.0\textwidth}{!}{
    \begin{tabular}{crrrrrrrrrrrrrrrrrrrrrr}
    \toprule
     \toprule
        \multicolumn{1}{c}{\textbf{Split}} & \multicolumn{1}{c|}{\textbf{Method}} & \multicolumn{1}{c}{\textbf{bike}} & \multicolumn{1}{c}{\textbf{build.}} & \multicolumn{1}{c}{\textbf{car}} & \multicolumn{1}{c}{\textbf{cone}} & \multicolumn{1}{c}{\textbf{fence}} & \multicolumn{1}{c}{\textbf{grou.}} & \multicolumn{1}{c}{\textbf{pers.}} & \multicolumn{1}{c}{\textbf{plants}} & \multicolumn{1}{c}{\textbf{pole}} & \multicolumn{1}{c}{\textbf{rider}} & \multicolumn{1}{c}{\textbf{traf.}} & \multicolumn{1}{c}{\textbf{trashc.}} & \multicolumn{1}{c|}{\textbf{trunk}} & \multicolumn{1}{c}{\textbf{Novel}} & \multicolumn{1}{c}{\textbf{Known}} & \multicolumn{1}{c}{\textbf{All}} \\
    \midrule
          & \multicolumn{1}{c|}{Full} & \multicolumn{1}{c}{46.9 } & \multicolumn{1}{c}{85.9 } & \multicolumn{1}{c}{55.4 } & \multicolumn{1}{c}{37.3 } & \multicolumn{1}{c}{48.9 } & \multicolumn{1}{c}{80.0} & \multicolumn{1}{c}{64.5 } & \multicolumn{1}{c}{79.4 } & \multicolumn{1}{c}{35.8 } & \multicolumn{1}{c}{59.4 } & \multicolumn{1}{c}{31.9 } & \multicolumn{1}{c}{7.8 } & \multicolumn{1}{c|}{25.2 } & \multicolumn{1}{c}{-} & \multicolumn{1}{c}{-} & \multicolumn{1}{c}{50.6 } \\
    \midrule
    \multicolumn{1}{c}{\multirow{4}[2]{*}{0}} & \multicolumn{1}{c|}{EUMS\cite{EUMS}} & \multicolumn{1}{c}{25.7 } & \multicolumn{1}{c}{\cellcolor{customcolor}4.0 } & \multicolumn{1}{c}{\cellcolor{customcolor}0.6 } & \multicolumn{1}{c}{16.4 } & \multicolumn{1}{c}{29.4 } & \multicolumn{1}{c}{\cellcolor{customcolor}36.8 } & \multicolumn{1}{c}{43.8 } & \multicolumn{1}{c}{\cellcolor{customcolor}28.5 } & \multicolumn{1}{c}{13.1 } & \multicolumn{1}{c}{26.8 } & \multicolumn{1}{c}{18.2 } & \multicolumn{1}{c}{3.3 } & \multicolumn{1}{c|}{16.9 } & \multicolumn{1}{c}{\cellcolor{customcolor}17.4 } & \multicolumn{1}{c}{21.5 } & \multicolumn{1}{c}{20.3 } \\
          & \multicolumn{1}{c|}{NOPS\cite{NOPS}} & \multicolumn{1}{c}{35.5 } & \multicolumn{1}{c}{\cellcolor{customcolor}30.4 } & \multicolumn{1}{c}{\cellcolor{customcolor}1.2 } & \multicolumn{1}{c}{13.5 } & \multicolumn{1}{c}{24.1 } & \multicolumn{1}{c}{\cellcolor{customcolor}69.1 } & \multicolumn{1}{c}{44.7 } & \multicolumn{1}{c}{\cellcolor{customcolor}42.1 } & \multicolumn{1}{c}{19.2 } & \multicolumn{1}{c}{47.7 } & \multicolumn{1}{c}{24.4 } & \multicolumn{1}{c}{8.2 } & \multicolumn{1}{c|}{21.8 } & \multicolumn{1}{c}{\cellcolor{customcolor}35.7 } & \multicolumn{1}{c}{26.6 } & \multicolumn{1}{c}{29.4 } \\
          & \multicolumn{1}{c|}{SNOPS \cite{SNOPS}} & \multicolumn{1}{c}{34.2 } & \multicolumn{1}{c}{\cellcolor{customcolor}\textbf{58.8} } & \multicolumn{1}{c}{\cellcolor{customcolor}\textbf{10.0} } & \multicolumn{1}{c}{13.2 } & \multicolumn{1}{c}{18.7 } & \multicolumn{1}{c}{\cellcolor{customcolor}77.3 } & \multicolumn{1}{c}{45.8 } & \multicolumn{1}{c}{\cellcolor{customcolor}\textbf{58.6} } & \multicolumn{1}{c}{17.3 } & \multicolumn{1}{c}{48.4 } & \multicolumn{1}{c}{22.6 } & \multicolumn{1}{c}{8.7 } & \multicolumn{1}{c|}{22.9 } & \multicolumn{1}{c}{\cellcolor{customcolor}\textbf{51.2} } & \multicolumn{1}{c}{25.8 } & \multicolumn{1}{c}{33.6 } \\
          & \multicolumn{1}{c|}{DASL\cite{DLAS}} & \multicolumn{1}{c}{46.3 } & \multicolumn{1}{c}{\cellcolor{customcolor}51.5 } & \multicolumn{1}{c}{\cellcolor{customcolor}6.0 } & \multicolumn{1}{c}{35.7 } & \multicolumn{1}{c}{48.5 } & \multicolumn{1}{c}{\cellcolor{customcolor}\textbf{83.0} } & \multicolumn{1}{c}{67.9 } & \multicolumn{1}{c}{\cellcolor{customcolor}53.1 } & \multicolumn{1}{c}{35.5 } & \multicolumn{1}{c}{59.3 } & \multicolumn{1}{c}{31.0 } & \multicolumn{1}{c}{2.8 } & \multicolumn{1}{c|}{15.5 } & \multicolumn{1}{c}{\cellcolor{customcolor}48.4 } & \multicolumn{1}{c}{38.0 } & \multicolumn{1}{c}{41.2} \\
          & \multicolumn{1}{c|}{Ours} & \multicolumn{1}{c}{44.7 } & \multicolumn{1}{c}{\cellcolor{customcolor}58.3 } & \multicolumn{1}{c}{\cellcolor{customcolor}4.4 } & \multicolumn{1}{c}{28.4 } & \multicolumn{1}{c}{47.4 } & \multicolumn{1}{c}{\cellcolor{customcolor}80.9 } & \multicolumn{1}{c}{65.4 } & \multicolumn{1}{c}{\cellcolor{customcolor}45.3 } & \multicolumn{1}{c}{41.6 } & \multicolumn{1}{c}{57.9 } & \multicolumn{1}{c}{34.1 } & \multicolumn{1}{c}{10.8 } & \multicolumn{1}{c|}{17.9 } & \multicolumn{1}{c}{\cellcolor{customcolor}47.2 } & \multicolumn{1}{c}{\textbf{38.7} } & \multicolumn{1}{c}{\textbf{41.3} } \\
    \midrule
    \multicolumn{1}{c}{\multirow{4}[2]{*}{1}} & \multicolumn{1}{c|}{EUMS\cite{EUMS}} & \multicolumn{1}{c}{\cellcolor{customcolor}15.2 } & \multicolumn{1}{c}{68.0 } & \multicolumn{1}{c}{28.0 } & \multicolumn{1}{c}{24.0 } & \multicolumn{1}{c}{\cellcolor{customcolor}11.9 } & \multicolumn{1}{c}{75.1 } & \multicolumn{1}{c}{\cellcolor{customcolor}36.0 } & \multicolumn{1}{c}{74.5 } & \multicolumn{1}{c}{26.9 } & \multicolumn{1}{c}{48.6 } & \multicolumn{1}{c}{26.0 } & \multicolumn{1}{c}{5.6 } & \multicolumn{1}{c|}{23.1 } & \multicolumn{1}{c}{\cellcolor{customcolor}21.0 } & \multicolumn{1}{c}{40.0 } & \multicolumn{1}{c}{35.6 } \\
          & \multicolumn{1}{c|}{NOPS\cite{NOPS}} & \multicolumn{1}{c}{\cellcolor{customcolor}29.4 } & \multicolumn{1}{c}{71.4 } & \multicolumn{1}{c}{28.7 } & \multicolumn{1}{c}{12.2 } & \multicolumn{1}{c}{\cellcolor{customcolor}3.9 } & \multicolumn{1}{c}{78.2 } & \multicolumn{1}{c}{\cellcolor{customcolor}\textbf{56.8} } & \multicolumn{1}{c}{74.2 } & \multicolumn{1}{c}{18.3 } & \multicolumn{1}{c}{38.9 } & \multicolumn{1}{c}{23.3 } & \multicolumn{1}{c}{13.7 } & \multicolumn{1}{c|}{23.5 } & \multicolumn{1}{c}{\cellcolor{customcolor}30.0 } & \multicolumn{1}{c}{38.2 } & \multicolumn{1}{c}{36.4 } \\
           & \multicolumn{1}{c|}{SNOPS \cite{SNOPS}} & \multicolumn{1}{c}{\cellcolor{customcolor}16.3 } & \multicolumn{1}{c}{71.4 } & \multicolumn{1}{c}{30.0 } & \multicolumn{1}{c}{19.8 } & \multicolumn{1}{c}{\cellcolor{customcolor}24.9 } & \multicolumn{1}{c}{77.1 } & \multicolumn{1}{c}{\cellcolor{customcolor}55.0 } & \multicolumn{1}{c}{73.4 } & \multicolumn{1}{c}{15.8 } & \multicolumn{1}{c}{38.4 } & \multicolumn{1}{c}{22.3 } & \multicolumn{1}{c}{15.7 } & \multicolumn{1}{c|}{23.6 } & \multicolumn{1}{c}{\cellcolor{customcolor}32.1 } & \multicolumn{1}{c}{38.7 } & \multicolumn{1}{c}{37.2 } \\
          & \multicolumn{1}{c|}{DASL\cite{DLAS}} & \multicolumn{1}{c}{\cellcolor{customcolor}31.5 } & \multicolumn{1}{c}{83.2 } & \multicolumn{1}{c}{48.7 } & \multicolumn{1}{c}{25.4 } & \multicolumn{1}{c}{\cellcolor{customcolor}23.9 } & \multicolumn{1}{c}{77.3 } & \multicolumn{1}{c}{\cellcolor{customcolor}53.1 } & \multicolumn{1}{c}{77.1 } & \multicolumn{1}{c}{32.5 } & \multicolumn{1}{c}{57.3 } & \multicolumn{1}{c}{35.0 } & \multicolumn{1}{c}{9.3 } & \multicolumn{1}{c|}{18.0 } & \multicolumn{1}{c}{\cellcolor{customcolor}36.2 } & \multicolumn{1}{c}{46.4 } & \multicolumn{1}{c}{44.0 } \\
          & \multicolumn{1}{c|}{Ours} & \multicolumn{1}{c}{\cellcolor{customcolor}\textbf{36.9} } & \multicolumn{1}{c}{83.9 } & \multicolumn{1}{c}{46.7 } & \multicolumn{1}{c}{26.7 } & \multicolumn{1}{c}{\cellcolor{customcolor}\textbf{29.6} } & \multicolumn{1}{c}{78.5 } & \multicolumn{1}{c}{\cellcolor{customcolor}49.1 } & \multicolumn{1}{c}{77.1 } & \multicolumn{1}{c}{36.2 } & \multicolumn{1}{c}{58.6 } & \multicolumn{1}{c}{33.4 } & \multicolumn{1}{c}{13.0 } & \multicolumn{1}{c|}{16.7 } & \multicolumn{1}{c}{\cellcolor{customcolor}\textbf{38.5} } & \multicolumn{1}{c}{\textbf{47.1} } & \multicolumn{1}{c}{\textbf{45.1} } \\
    \midrule
    \multicolumn{1}{c}{\multirow{4}[2]{*}{2}} & \multicolumn{1}{c|}{EUMS\cite{EUMS}} & \multicolumn{1}{c}{40.1 } & \multicolumn{1}{c}{69.5 } & \multicolumn{1}{c}{27.7 } & \multicolumn{1}{c}{13.5 } & \multicolumn{1}{c}{34.9 } & \multicolumn{1}{c}{76.0 } & \multicolumn{1}{c}{54.7 } & \multicolumn{1}{c}{75.6 } & \multicolumn{1}{c}{\cellcolor{customcolor}5.3 } & \multicolumn{1}{c}{39.2 } & \multicolumn{1}{c}{\cellcolor{customcolor}7.8 } & \multicolumn{1}{c}{8.5 } & \multicolumn{1}{c|}{\cellcolor{customcolor}11.9 } & \multicolumn{1}{c}{\cellcolor{customcolor}8.3 } & \multicolumn{1}{c}{44.0 } & \multicolumn{1}{c}{35.7 } \\
          & \multicolumn{1}{c|}{NOPS\cite{NOPS}} & \multicolumn{1}{c}{37.2 } & \multicolumn{1}{c}{71.8 } & \multicolumn{1}{c}{29.7 } & \multicolumn{1}{c}{14.6 } & \multicolumn{1}{c}{28.4 } & \multicolumn{1}{c}{77.5 } & \multicolumn{1}{c}{52.1 } & \multicolumn{1}{c}{73.0 } & \multicolumn{1}{c}{\cellcolor{customcolor}11.5 } & \multicolumn{1}{c}{47.1 } & \multicolumn{1}{c}{\cellcolor{customcolor}0.5 } & \multicolumn{1}{c}{10.2 } & \multicolumn{1}{c|}{\cellcolor{customcolor}14.8 } & \multicolumn{1}{c}{\cellcolor{customcolor}9.0 } & \multicolumn{1}{c}{44.2 } & \multicolumn{1}{c}{36.0 } \\
          & \multicolumn{1}{c|}{SNOPS \cite{SNOPS}} & \multicolumn{1}{c}{38.4 } & \multicolumn{1}{c}{72.5 } & \multicolumn{1}{c}{28.0 } & \multicolumn{1}{c}{14.5 } & \multicolumn{1}{c}{26.2 } & \multicolumn{1}{c}{78.1 } & \multicolumn{1}{c}{54.7 } & \multicolumn{1}{c}{74.3 } & \multicolumn{1}{c}{\cellcolor{customcolor}10.0 } & \multicolumn{1}{c}{48.3 } & \multicolumn{1}{c}{\cellcolor{customcolor}23.0 } & \multicolumn{1}{c}{10.2 } & \multicolumn{1}{c|}{\cellcolor{customcolor}\textbf{17.7} } & \multicolumn{1}{c}{\cellcolor{customcolor}16.9 } & \multicolumn{1}{c}{44.5 } & \multicolumn{1}{c}{38.1 } \\
          & \multicolumn{1}{c|}{DASL\cite{DLAS}} & \multicolumn{1}{c}{45.3 } & \multicolumn{1}{c}{82.8 } & \multicolumn{1}{c}{49.8 } & \multicolumn{1}{c}{28.4 } & \multicolumn{1}{c}{46.3 } & \multicolumn{1}{c}{76.7 } & \multicolumn{1}{c}{66.2 } & \multicolumn{1}{c}{77.2 } & \multicolumn{1}{c}{\cellcolor{customcolor}10.9 } & \multicolumn{1}{c}{58.4 } & \multicolumn{1}{c}{\cellcolor{customcolor}18.6 } & \multicolumn{1}{c}{7.3} & \multicolumn{1}{c|}{\cellcolor{customcolor}8.2 } & \multicolumn{1}{c}{\cellcolor{customcolor}12.6 } & \multicolumn{1}{c}{53.8 } & \multicolumn{1}{c}{44.3 } \\
          & \multicolumn{1}{c|}{Ours} & \multicolumn{1}{c}{45.5 } & \multicolumn{1}{c}{82.5 } & \multicolumn{1}{c}{49.2 } & \multicolumn{1}{c}{37.3 } & \multicolumn{1}{c}{43.3 } & \multicolumn{1}{c}{77.8 } & \multicolumn{1}{c}{67.7 } & \multicolumn{1}{c}{78.0 } & \multicolumn{1}{c}{\cellcolor{customcolor}\textbf{21.7} } & \multicolumn{1}{c}{59.4 } & \multicolumn{1}{c}{\cellcolor{customcolor}\textbf{36.0} } & \multicolumn{1}{c}{3.3} & \multicolumn{1}{c|}{\cellcolor{customcolor}9.3 } & \multicolumn{1}{c}{\cellcolor{customcolor}\textbf{22.3} } & \multicolumn{1}{c}{\textbf{54.4} } & \multicolumn{1}{c}{\textbf{47.0} } \\
    \midrule
    \multicolumn{1}{c}{\multirow{4}[2]{*}{3}} & \multicolumn{1}{c|}{EUMS\cite{EUMS}} & \multicolumn{1}{c}{41.2 } & \multicolumn{1}{c}{70.7 } & \multicolumn{1}{c}{28.1 } & \multicolumn{1}{c}{\cellcolor{customcolor}4.3 } & \multicolumn{1}{c}{38.3 } & \multicolumn{1}{c}{76.7 } & \multicolumn{1}{c}{38.3 } & \multicolumn{1}{c}{75.4 } & \multicolumn{1}{c}{25.8 } & \multicolumn{1}{c}{\cellcolor{customcolor}34.3 } & \multicolumn{1}{c}{28.3 } & \multicolumn{1}{c}{\cellcolor{customcolor}0.4 } & \multicolumn{1}{c|}{24.4 } & \multicolumn{1}{c}{\cellcolor{customcolor}13.0 } & \multicolumn{1}{c}{44.7 } & \multicolumn{1}{c}{37.4 } \\
          & \multicolumn{1}{c|}{NOPS\cite{NOPS}} & \multicolumn{1}{c}{38.6 } & \multicolumn{1}{c}{70.4 } & \multicolumn{1}{c}{30.9 } & \multicolumn{1}{c}{\cellcolor{customcolor}0.0 } & \multicolumn{1}{c}{32.8 } & \multicolumn{1}{c}{76.5 } & \multicolumn{1}{c}{50.6 } & \multicolumn{1}{c}{71.8 } & \multicolumn{1}{c}{17.0 } & \multicolumn{1}{c}{\cellcolor{customcolor}31.9 } & \multicolumn{1}{c}{36.3 } & \multicolumn{1}{c}{\cellcolor{customcolor}1.0 } & \multicolumn{1}{c|}{22.6 } & \multicolumn{1}{c}{\cellcolor{customcolor}10.9 } & \multicolumn{1}{c}{43.9 } & \multicolumn{1}{c}{36.3 } \\
           & \multicolumn{1}{c|}{SNOPS \cite{SNOPS}} & \multicolumn{1}{c}{39.4 } & \multicolumn{1}{c}{70.3 } & \multicolumn{1}{c}{30.0 } & \multicolumn{1}{c}{\cellcolor{customcolor}9.1 } & \multicolumn{1}{c}{26.8 } & \multicolumn{1}{c}{77.6 } & \multicolumn{1}{c}{54.3 } & \multicolumn{1}{c}{72.5 } & \multicolumn{1}{c}{16.0 } & \multicolumn{1}{c}{\cellcolor{customcolor}49.9 } & \multicolumn{1}{c}{28.1 } & \multicolumn{1}{c}{\cellcolor{customcolor}1.3 } & \multicolumn{1}{c|}{23.5 } & \multicolumn{1}{c}{\cellcolor{customcolor}20.1 } & \multicolumn{1}{c}{43.9 } & \multicolumn{1}{c}{38.4 } \\
          & \multicolumn{1}{c|}{DASL\cite{DLAS}} & \multicolumn{1}{c}{45.5 } & \multicolumn{1}{c}{82.9 } & \multicolumn{1}{c}{47.7 } & \multicolumn{1}{c}{\cellcolor{customcolor}0.0 } & \multicolumn{1}{c}{45.1 } & \multicolumn{1}{c}{77.8 } & \multicolumn{1}{c}{66.3 } & \multicolumn{1}{c}{77.7 } & \multicolumn{1}{c}{34.3 } & \multicolumn{1}{c}{\cellcolor{customcolor}49.1 } & \multicolumn{1}{c}{30.5 } & \multicolumn{1}{c}{\cellcolor{customcolor}\textbf{4.0} } & \multicolumn{1}{c|}{15.3 } & \multicolumn{1}{c}{\cellcolor{customcolor}17.7 } & \multicolumn{1}{c}{\textbf{52.8} } & \multicolumn{1}{c}{44.7 } \\
          & \multicolumn{1}{c|}{Ours} & \multicolumn{1}{c}{44.2 } & \multicolumn{1}{c}{83.0 } & \multicolumn{1}{c}{50.2 } & \multicolumn{1}{c}{\cellcolor{customcolor}\textbf{25.7} } & \multicolumn{1}{c}{46.9 } & \multicolumn{1}{c}{77.4 } & \multicolumn{1}{c}{67.0 } & \multicolumn{1}{c}{77.5 } & \multicolumn{1}{c}{35.6 } & \multicolumn{1}{c}{\cellcolor{customcolor}\textbf{51.3} } & \multicolumn{1}{c}{36.5 } & \multicolumn{1}{c}{\cellcolor{customcolor}0.0 } & \multicolumn{1}{c|}{13.5 } & \multicolumn{1}{c}{\cellcolor{customcolor}\textbf{37.8} } & \multicolumn{1}{c}{49.5 } & \multicolumn{1}{c}{\textbf{46.8} } \\
      \bottomrule
      \bottomrule
    \end{tabular}
    
  }
  \vspace{-10pt}
  \label{tab1}
\end{table*}


\subsection{Further Discussion}
In this section, we further discuss the innovation and applicability of the proposed method of Geometric-Aware Hypergraph Reasoning in point cloud segmentation and Novel Class Discovery (NCD) tasks.

To the best of our knowledge, this is the first work introducing high-order hypergraph reasoning into the NCD task. By connecting multiple prototype nodes through hyperedges, our method captures complex inter-category and part-level relationships beyond the binary constraints of conventional graphs. This design naturally aligns with the sparsity and geometric nature of point clouds, enabling joint semantic and geometric knowledge transfer and improving the discriminability and generalization of unseen categories. Moreover, the dynamic hyperedge adjustment mechanism adaptively updates the connectivity and weights according to semantic and geometric similarities, effectively addressing novel class emergence and class imbalance in the open-world scenarios. Extensive experimental results and visualization analysis demonstrate the superiorities of our method.

Finally, compared with open-vocabulary methods \cite{uni3d, OVD2}, NCD exhibits distinct advantages. Open-vocabulary approaches rely heavily on the priors of large language models, resulting in the performance degradation when encountering concepts beyond their semantic coverage, and they are often unsuitable for text-sparse 3D environments. In contrast, NCD infers novel categories through self-organized reasoning over known classes without linguistic priors, making it inherently more compatible with purely visual and geometric understanding tasks. In the supplementary material, based on the same settings, our method still outperforms open-vocabulary methods, indicating the effectiveness.

\section{Experiments}
\vspace{-5pt}
In the experiments, for Novel Class Discovery in Point Cloud Segmentation, we adopt the settings of NOPS \cite{NOPS} and DASL \cite{DLAS} to evaluate the model's ability to segment base and novel classes.
\vspace{-5pt}
\subsection{Experimental setup}

\textbf{Dataset.} We evaluated our approach on the widely used SemanticKITTI \cite{KITTI} and SemanticPOSS \cite{POSS} datasets, with 19 and 13 semantic classes, respectively. We adopt the same dataset splitting strategy for a fair comparison with existing methods \cite{DLAS, NOPS}, where one subset is selected as novel classes and the others as known classes. 

\textbf{Evaluation Metric.} 
Following the dataset splits from previous works \cite{NOPS, DLAS} on the SemanticKITTI \cite{KITTI, K2, kitti2} and SemanticPOSS \cite{POSS} datasets, we conducted evaluations on sequences 08 and 03, respectively. These sequences contain both known and novel classes. In each split, novel classes correspond to unlabeled points, while base classes correspond to labeled points. The splits are based on class distribution and the semantic relationships between novel and base classes. We report the IoU for known classes, compute the IoU for novel classes after Hungarian matching, and further take the column-wise mean over all classes to assess overall performance. Further details can be found in the Supplementary Material.



\vspace{-8pt}
\subsection{Implementation Details}
\vspace{-5pt}
To ensure a fair comparison, we implemented our network using the same MinkowskiUNet-34C \cite{Mink} architecture as the other two works\cite{NOPS, DLAS}, with point-level features extracted from the second-to-last layer. The segmentation head is realized through a linear layer, generating output logits for each point in the point cloud for every batch. The optimizer is AdamW \cite{AdamW}, with an initial learning rate of 1e-3, which decays every 5 epochs, ultimately reaching a minimum value of 1e-5. Regarding hyperparameters, the number of geometric-aware prototypes \( K \) is consistent with the number of categories in the dataset, and the weighting coefficient \( \alpha \) for geometric and semantic similarity starts at 0.5, adjusting dynamically during network training. The hyperparameter \( M \), representing the number of connections for each prototype with others, affects the complexity of the hypergraph structure. Ablation experiments confirm that the best performance is achieved when \( M = 8 \). The smoothing parameter of the label \( \epsilon \) is set to 0.15 to reduce overfitting during pseudo-label generation.

\vspace{-8pt}
\subsection{Comparison with the State of the Art}
\vspace{-5pt}
\textbf{SemanticPOSS Dataset.}
We compared our method with the current state-of-the-art. As shown in Table \ref{tab1}, our method consistently outperforms others across all four splits of the SemanticPOSS Dataset, with significant improvements in novel class segmentation on Splits 2 and 3. Specifically, we observe substantial gains ranging from 16.9 to 22.3 on Split 2 and from 20.1 to 37.8 on Split 3, nearly doubling the performance. The results demonstrate that the proposed method effectively improves segmentation performance on novel classes by leveraging hypergraph-based reasoning. We also include segmentation comparisons for indoor scenes and additional ablation results for small objects; details are provided in the Supplementary Material.

\begin{table*}  
  \centering
  \vspace{-20pt}
  \caption{The novel class discovery results on the SemanticKITTI dataset. ‘Full’ denotes
the results obtained by supervised learning. The four groups represent the four splits in
turn, and the gray values are the novel classes in each split.}
  \fontsize{22}{23}\selectfont
     \resizebox{\textwidth}{!}{
    \begin{tabular}{crrrrrrrrrrrrrrrrrrrrrr}
    \toprule
    \toprule
    \multicolumn{1}{c|}{\textbf{Method}} & \multicolumn{1}{c}{\textbf{bi.cle}} & \multicolumn{1}{c}{\textbf{b.clst}} & \multicolumn{1}{c}{\textbf{build.}} & \multicolumn{1}{c}{\textbf{car}} & \multicolumn{1}{c}{\textbf{fence}} & \multicolumn{1}{c}{\textbf{mt.cle}} & \multicolumn{1}{c}{\textbf{m.clst}} & \multicolumn{1}{c}{\textbf{oth-g.}} & \multicolumn{1}{c}{\textbf{oth-v.}} & \multicolumn{1}{c}{\textbf{park.}} & \multicolumn{1}{c}{\textbf{pers.}} & \multicolumn{1}{c}{\textbf{pole}} & \multicolumn{1}{c}{\textbf{road}} & \multicolumn{1}{c}{\textbf{sidew.}} & \multicolumn{1}{c}{\textbf{terra.}} & \multicolumn{1}{c}{\textbf{traff.}} & \multicolumn{1}{c}{\textbf{truck}} & \multicolumn{1}{c}{\textbf{trunk}} & \multicolumn{1}{c|}{\textbf{veget.}} & \multicolumn{1}{c}{\textbf{Novel}} & \multicolumn{1}{c}{\textbf{Known}} & \multicolumn{1}{c}{\textbf{All}} \\
    \midrule
    \multicolumn{1}{c|}{Full} & \multicolumn{1}{c}{3.6 } & \multicolumn{1}{c}{58.3 } & \multicolumn{1}{c}{88.5 } & \multicolumn{1}{c}{93.4 } & \multicolumn{1}{c}{36.9 } & \multicolumn{1}{c}{25.4 } & \multicolumn{1}{c}{0.9 } & \multicolumn{1}{c}{2.0 } & \multicolumn{1}{c}{26.9 } & \multicolumn{1}{c}{37.8 } & \multicolumn{1}{c}{36.2 } & \multicolumn{1}{c}{62.6 } & \multicolumn{1}{c}{92.5 } & \multicolumn{1}{c}{76.8 } & \multicolumn{1}{c}{62.5 } & \multicolumn{1}{c}{40.4 } & \multicolumn{1}{c}{55.7 } & \multicolumn{1}{c}{58.4} & \multicolumn{1}{c|}{89.0 } & \multicolumn{1}{c}{- } & \multicolumn{1}{c}{- }        & \multicolumn{1}{c}{49.8 } \\
    \midrule
    \multicolumn{1}{c|}{EUMS\cite{EUMS}} & \multicolumn{1}{c}{5.3 } & \multicolumn{1}{c}{40.0 } & \multicolumn{1}{c}{\cellcolor{customcolor}15.8 } & \multicolumn{1}{c}{79.2 } & \multicolumn{1}{c}{9.0 } & \multicolumn{1}{c}{16.9 } & \multicolumn{1}{c}{2.5 } & \multicolumn{1}{c}{0.1 } & \multicolumn{1}{c}{11.4 } & \multicolumn{1}{c}{14.4 } & \multicolumn{1}{c}{12.7 } & \multicolumn{1}{c}{29.2 } & \multicolumn{1}{c}{\cellcolor{customcolor}42.6 } & \multicolumn{1}{c}{\cellcolor{customcolor}26.1 } & \multicolumn{1}{c}{\cellcolor{customcolor}0.1 } & \multicolumn{1}{c}{10.3 } & \multicolumn{1}{c}{47.4 } & \multicolumn{1}{c}{37.9 } & \multicolumn{1}{c|}{\cellcolor{customcolor}38.4 } & \multicolumn{1}{c}{\cellcolor{customcolor}24.6 } & \multicolumn{1}{c}{21.1 } & \multicolumn{1}{c}{23.1 } \\
    \multicolumn{1}{c|}{NOPS\cite{NOPS}} & \multicolumn{1}{c}{5.6 } & \multicolumn{1}{c}{47.8 } & \multicolumn{1}{c}{\cellcolor{customcolor}52.7 } & \multicolumn{1}{c}{82.6 } & \multicolumn{1}{c}{13.8 } & \multicolumn{1}{c}{25.6 } & \multicolumn{1}{c}{1.4 } & \multicolumn{1}{c}{1.7 } & \multicolumn{1}{c}{14.5 } & \multicolumn{1}{c}{19.8 } & \multicolumn{1}{c}{25.9 } & \multicolumn{1}{c}{32.1 } & \multicolumn{1}{c}{\cellcolor{customcolor}\textbf{56.7} } & \multicolumn{1}{c}{\cellcolor{customcolor}8.1 } & \multicolumn{1}{c}{\cellcolor{customcolor}23.8 } & \multicolumn{1}{c}{14.3 } & \multicolumn{1}{c}{49.4 } & \multicolumn{1}{c}{36.2 } & \multicolumn{1}{c|}{\cellcolor{customcolor}44.2 } & \multicolumn{1}{c}{\cellcolor{customcolor}37.1 } & \multicolumn{1}{c}{26.5 } & \multicolumn{1}{c}{29.3 } \\
    \multicolumn{1}{c|}{SNOPS \cite{SNOPS}} & \multicolumn{1}{c}{6.6 } & \multicolumn{1}{c}{43.9 } & \multicolumn{1}{c}{\cellcolor{customcolor}72.0 } & \multicolumn{1}{c}{83.3 } & \multicolumn{1}{c}{13.6 } & \multicolumn{1}{c}{24.7 } & \multicolumn{1}{c}{2.5 } & \multicolumn{1}{c}{2.4 } & \multicolumn{1}{c}{15.1 } & \multicolumn{1}{c}{18.7 } & \multicolumn{1}{c}{24.6 } & \multicolumn{1}{c}{31.6 } & \multicolumn{1}{c}{\cellcolor{customcolor}49.5 } & \multicolumn{1}{c}{\cellcolor{customcolor}\textbf{43.2} } & \multicolumn{1}{c}{\cellcolor{customcolor}\textbf{27.4} } & \multicolumn{1}{c}{15.7 } & \multicolumn{1}{c}{42.1 } & \multicolumn{1}{c}{38.5 } & \multicolumn{1}{c|}{\cellcolor{customcolor}37.5 } & \multicolumn{1}{c}{\cellcolor{customcolor}45.9 } & \multicolumn{1}{c}{26.0 } & \multicolumn{1}{c}{31.2 } \\
    \multicolumn{1}{c|}{DASL\cite{DLAS}} & \multicolumn{1}{c}{5.5 } & \multicolumn{1}{c}{51.1 } & \multicolumn{1}{c}{\cellcolor{customcolor}\textbf{74.6} } & \multicolumn{1}{c}{92.3 } & \multicolumn{1}{c}{29.8 } & \multicolumn{1}{c}{22.8 } & \multicolumn{1}{c}{0.0 } & \multicolumn{1}{c}{0.0 } & \multicolumn{1}{c}{23.3 } & \multicolumn{1}{c}{24.8 } & \multicolumn{1}{c}{27.7 } & \multicolumn{1}{c}{59.7 } & \multicolumn{1}{c}{\cellcolor{customcolor}41.4 } & \multicolumn{1}{c}{\cellcolor{customcolor}22.5 } & \multicolumn{1}{c}{\cellcolor{customcolor}23.6 } & \multicolumn{1}{c}{39.3 } & \multicolumn{1}{c}{43.6 } & \multicolumn{1}{c}{51.1 } & \multicolumn{1}{c|}{\cellcolor{customcolor}\textbf{66.4} } & \multicolumn{1}{c}{\cellcolor{customcolor}45.7 } & \multicolumn{1}{c}{\textbf{33.7} } & \multicolumn{1}{c}{\textbf{36.8} } \\
    \multicolumn{1}{c|}{Ours} & \multicolumn{1}{c}{5.1 } & \multicolumn{1}{c}{48.4 } & \multicolumn{1}{c}{\cellcolor{customcolor}69.3 } & \multicolumn{1}{c}{92.5 } & \multicolumn{1}{c}{29.6 } & \multicolumn{1}{c}{23.1 } & \multicolumn{1}{c}{0.3 } & \multicolumn{1}{c}{0.0 } & \multicolumn{1}{c}{25.3 } & \multicolumn{1}{c}{25.2 } & \multicolumn{1}{c}{22.3} & \multicolumn{1}{c}{59.4 } & \multicolumn{1}{c}{\cellcolor{customcolor}52.7 } & \multicolumn{1}{c}{\cellcolor{customcolor}26.3 } & \multicolumn{1}{c}{\cellcolor{customcolor}25.9 } & \multicolumn{1}{c}{36.3 } & \multicolumn{1}{c}{42.8 } & \multicolumn{1}{c}{51.6 } & \multicolumn{1}{c|}{\cellcolor{customcolor}63.4 } & \multicolumn{1}{c}{\cellcolor{customcolor}\textbf{47.5} } & \multicolumn{1}{c}{33.0 } & \multicolumn{1}{c}{\textbf{36.8} } \\
    \midrule
    \multicolumn{1}{c|}{EUMS\cite{EUMS}} & \multicolumn{1}{c}{7.5 } & \multicolumn{1}{c}{42.4 } & \multicolumn{1}{c}{80.0 } & \multicolumn{1}{c}{\cellcolor{customcolor}\textbf{76.8} } & \multicolumn{1}{c}{\cellcolor{customcolor}8.6 } & \multicolumn{1}{c}{19.6 } & \multicolumn{1}{c}{1.4 } & \multicolumn{1}{c}{\cellcolor{customcolor}\textbf{0.6} } & \multicolumn{1}{c}{12.0 } & \multicolumn{1}{c}{\cellcolor{customcolor}14.1 } & \multicolumn{1}{c}{14.0 } & \multicolumn{1}{c}{40.7 } & \multicolumn{1}{c}{86.3 } & \multicolumn{1}{c}{66.5 } & \multicolumn{1}{c}{56.3 } & \multicolumn{1}{c}{12.0 } & \multicolumn{1}{c}{44.8 } & \multicolumn{1}{c}{\cellcolor{customcolor}20.9 } & \multicolumn{1}{c|}{72.4 } & \multicolumn{1}{c}{\cellcolor{customcolor}24.2 } & \multicolumn{1}{c}{37.1 } & \multicolumn{1}{c}{35.6 } \\
    \multicolumn{1}{c|}{NOPS\cite{NOPS}} & \multicolumn{1}{c}{7.4 } & \multicolumn{1}{c}{51.2 } & \multicolumn{1}{c}{84.5 } & \multicolumn{1}{c}{\cellcolor{customcolor}50.9 } & \multicolumn{1}{c}{\cellcolor{customcolor}7.3 } & \multicolumn{1}{c}{28.9 } & \multicolumn{1}{c}{1.8 } & \multicolumn{1}{c}{\cellcolor{customcolor}0.0 } & \multicolumn{1}{c}{22.2 } & \multicolumn{1}{c}{\cellcolor{customcolor}19.4 } & \multicolumn{1}{c}{30.4 } & \multicolumn{1}{c}{37.6 } & \multicolumn{1}{c}{90.1 } & \multicolumn{1}{c}{72.2 } & \multicolumn{1}{c}{60.8 } & \multicolumn{1}{c}{16.8 } & \multicolumn{1}{c}{57.3 } & \multicolumn{1}{c}{\cellcolor{customcolor}\textbf{49.3} } & \multicolumn{1}{c|}{85.1 } & \multicolumn{1}{c}{\cellcolor{customcolor}25.4 } & \multicolumn{1}{c}{46.2 } & \multicolumn{1}{c}{40.7 } \\
     \multicolumn{1}{c|}{SNOPS \cite{SNOPS}} & \multicolumn{1}{c}{7.6 } & \multicolumn{1}{c}{43.5 } & \multicolumn{1}{c}{85.1 } & \multicolumn{1}{c}{\cellcolor{customcolor}68.7 } & \multicolumn{1}{c}{\cellcolor{customcolor}\textbf{18.9} } & \multicolumn{1}{c}{24.4 } & \multicolumn{1}{c}{3.5 } & \multicolumn{1}{c}{\cellcolor{customcolor}0.0 } & \multicolumn{1}{c}{23.9 } & \multicolumn{1}{c}{\cellcolor{customcolor}19.1 } & \multicolumn{1}{c}{27.0 } & \multicolumn{1}{c}{36.5 } & \multicolumn{1}{c}{89.3 } & \multicolumn{1}{c}{71.9 } & \multicolumn{1}{c}{62.0 } & \multicolumn{1}{c}{17.2 } & \multicolumn{1}{c}{55.9 } & \multicolumn{1}{c}{\cellcolor{customcolor}29.4 } & \multicolumn{1}{c|}{84.4 } & \multicolumn{1}{c}{\cellcolor{customcolor}27.2 } & \multicolumn{1}{c}{45.2 } & \multicolumn{1}{c}{40.4 } \\
    \multicolumn{1}{c|}{DASL\cite{DLAS}} & \multicolumn{1}{c}{3.7 } & \multicolumn{1}{c}{57.4 } & \multicolumn{1}{c}{89.2 } & \multicolumn{1}{c}{\cellcolor{customcolor}56.5 } & \multicolumn{1}{c}{\cellcolor{customcolor}17.3 } & \multicolumn{1}{c}{20.3 } & \multicolumn{1}{c}{0.0 } & \multicolumn{1}{c}{\cellcolor{customcolor}0.0 } & \multicolumn{1}{c}{20.0 } & \multicolumn{1}{c}{\cellcolor{customcolor}\textbf{30.6} } & \multicolumn{1}{c}{34.8 } & \multicolumn{1}{c}{60.6 } & \multicolumn{1}{c}{93.2 } & \multicolumn{1}{c}{77.6 } & \multicolumn{1}{c}{62.0 } & \multicolumn{1}{c}{38.7 } & \multicolumn{1}{c}{56.9 } & \multicolumn{1}{c}{\cellcolor{customcolor}39.2 } & \multicolumn{1}{c|}{86.7 } & \multicolumn{1}{c}{\cellcolor{customcolor}28.7 } & \multicolumn{1}{c}{50.1 } & \multicolumn{1}{c}{44.5 } \\
    \multicolumn{1}{c|}{Ours} & \multicolumn{1}{c}{5.3 } & \multicolumn{1}{c}{64.4 } & \multicolumn{1}{c}{88.6 } & \multicolumn{1}{c}{\cellcolor{customcolor}76.1 } & \multicolumn{1}{c}{\cellcolor{customcolor}17.1 } & \multicolumn{1}{c}{21.9 } & \multicolumn{1}{c}{0.0 } & \multicolumn{1}{c}{\cellcolor{customcolor}0.0 } & \multicolumn{1}{c}{26.1 } & \multicolumn{1}{c}{\cellcolor{customcolor}24.2 } & \multicolumn{1}{c}{38.5 } & \multicolumn{1}{c}{57.3 } & \multicolumn{1}{c}{93.0 } & \multicolumn{1}{c}{76.6 } & \multicolumn{1}{c}{61.1 } & \multicolumn{1}{c}{38.8 } & \multicolumn{1}{c}{62.3 } & \multicolumn{1}{c}{\cellcolor{customcolor}41.3 } & \multicolumn{1}{c|}{87.0 } & \multicolumn{1}{c}{\cellcolor{customcolor}\textbf{31.7} } & \multicolumn{1}{c}{\textbf{50.5} } & \multicolumn{1}{c}{\textbf{46.3} } \\
    \midrule
    \multicolumn{1}{c|}{EUMS\cite{EUMS}} & \multicolumn{1}{c}{8.3 } & \multicolumn{1}{c}{50.8 } & \multicolumn{1}{c}{83.0 } & \multicolumn{1}{c}{88.1 } & \multicolumn{1}{c}{17.9 } & \multicolumn{1}{c}{\cellcolor{customcolor}2.8 } & \multicolumn{1}{c}{2.3 } & \multicolumn{1}{c}{0.2 } & \multicolumn{1}{c}{\cellcolor{customcolor}3.2 } & \multicolumn{1}{c}{25.4 } & \multicolumn{1}{c}{25.0 } & \multicolumn{1}{c}{\cellcolor{customcolor}20.2 } & \multicolumn{1}{c}{88.3 } & \multicolumn{1}{c}{71.0 } & \multicolumn{1}{c}{57.9 } & \multicolumn{1}{c}{\cellcolor{customcolor}8.6 } & \multicolumn{1}{c}{\cellcolor{customcolor}27.2 } & \multicolumn{1}{c}{38.4 } & \multicolumn{1}{c|}{77.0 } & \multicolumn{1}{c}{\cellcolor{customcolor}12.4 } & \multicolumn{1}{c}{42.2 } & \multicolumn{1}{c}{36.6 } \\
    \multicolumn{1}{c|}{NOPS\cite{NOPS}} & \multicolumn{1}{c}{6.7 } & \multicolumn{1}{c}{49.2 } & \multicolumn{1}{c}{86.4 } & \multicolumn{1}{c}{90.8 } & \multicolumn{1}{c}{23.7 } & \multicolumn{1}{c}{\cellcolor{customcolor}2.7 } & \multicolumn{1}{c}{0.6 } & \multicolumn{1}{c}{1.9 } & \multicolumn{1}{c}{\cellcolor{customcolor}\textbf{15.5} } & \multicolumn{1}{c}{29.5 } & \multicolumn{1}{c}{27.9 } & \multicolumn{1}{c}{\cellcolor{customcolor}36.4 } & \multicolumn{1}{c}{90.3 } & \multicolumn{1}{c}{73.4 } & \multicolumn{1}{c}{61.2 } & \multicolumn{1}{c}{\cellcolor{customcolor}17.8 } & \multicolumn{1}{c}{\cellcolor{customcolor}10.3 } & \multicolumn{1}{c}{46.2 } & \multicolumn{1}{c|}{84.3 } & \multicolumn{1}{c}{\cellcolor{customcolor}16.5 } & \multicolumn{1}{c}{48.0 } & \multicolumn{1}{c}{39.7 } \\
    \multicolumn{1}{c|}{SNOPS \cite{SNOPS}} & \multicolumn{1}{c}{6.8 } & \multicolumn{1}{c}{48.3 } & \multicolumn{1}{c}{86.1 } & \multicolumn{1}{c}{89.9 } & \multicolumn{1}{c}{22.2 } & \multicolumn{1}{c}{\cellcolor{customcolor}9.3 } & \multicolumn{1}{c}{0.6 } & \multicolumn{1}{c}{3.6 } & \multicolumn{1}{c}{\cellcolor{customcolor}10.5 } & \multicolumn{1}{c}{28.4 } & \multicolumn{1}{c}{27.1 } & \multicolumn{1}{c}{\cellcolor{customcolor}23.8 } & \multicolumn{1}{c}{90.6 } & \multicolumn{1}{c}{73.8 } & \multicolumn{1}{c}{61.9 } & \multicolumn{1}{c}{\cellcolor{customcolor}\textbf{22.3} } & \multicolumn{1}{c}{\cellcolor{customcolor}22.1 } & \multicolumn{1}{c}{46.1 } & \multicolumn{1}{c|}{83.8 } & \multicolumn{1}{c}{\cellcolor{customcolor}17.6 } & \multicolumn{1}{c}{48.0 } & \multicolumn{1}{c}{39.8 } \\
    \multicolumn{1}{c|}{DASL\cite{DLAS}} & \multicolumn{1}{c}{3.6 } & \multicolumn{1}{c}{54.2 } & \multicolumn{1}{c}{88.9 } & \multicolumn{1}{c}{93.3 } & \multicolumn{1}{c}{28.4 } & \multicolumn{1}{c}{\cellcolor{customcolor}\textbf{10.2} } & \multicolumn{1}{c}{0.0 } & \multicolumn{1}{c}{0.9 } & \multicolumn{1}{c}{\cellcolor{customcolor}9.6 } & \multicolumn{1}{c}{33.4 } & \multicolumn{1}{c}{32.2 } & \multicolumn{1}{c}{\cellcolor{customcolor}36.1 } & \multicolumn{1}{c}{92.7 } & \multicolumn{1}{c}{77.4 } & \multicolumn{1}{c}{62.2 } & \multicolumn{1}{c}{\cellcolor{customcolor}10.7 } & \multicolumn{1}{c}{\cellcolor{customcolor}\textbf{34.2} } & \multicolumn{1}{c}{51.7 } & \multicolumn{1}{c|}{86.9 } & \multicolumn{1}{c}{\cellcolor{customcolor}20.1 } & \multicolumn{1}{c}{50.4 } & \multicolumn{1}{c}{42.5 } \\
    \multicolumn{1}{c|}{Ours} & \multicolumn{1}{c}{5.7 } & \multicolumn{1}{c}{66.2 } & \multicolumn{1}{c}{89.0 } & \multicolumn{1}{c}{93.3 } & \multicolumn{1}{c}{27.6 } & \multicolumn{1}{c}{\cellcolor{customcolor}0.0 } & \multicolumn{1}{c}{0.0 } & \multicolumn{1}{c}{0.3 } & \multicolumn{1}{c}{\cellcolor{customcolor}14.3 } & \multicolumn{1}{c}{33.5 } & \multicolumn{1}{c}{37.3 } & \multicolumn{1}{c}{\cellcolor{customcolor}\textbf{51.1} } & \multicolumn{1}{c}{93.0 } & \multicolumn{1}{c}{77.0 } & \multicolumn{1}{c}{61.4 } & \multicolumn{1}{c}{\cellcolor{customcolor}9.4 } & \multicolumn{1}{c}{\cellcolor{customcolor}29.8 } & \multicolumn{1}{c}{54.2 } & \multicolumn{1}{c|}{86.7 } & \multicolumn{1}{c}{\cellcolor{customcolor}\textbf{20.9} } & \multicolumn{1}{c}{\textbf{51.8} } & \multicolumn{1}{c}{\textbf{43.7} } \\
    \midrule
    \multicolumn{1}{c|}{EUMS\cite{EUMS}} & \multicolumn{1}{c}{\cellcolor{customcolor}4.0 } & \multicolumn{1}{c}{\cellcolor{customcolor}2.5 } & \multicolumn{1}{c}{80.1 } & \multicolumn{1}{c}{87.2 } & \multicolumn{1}{c}{16.8 } & \multicolumn{1}{c}{14.0 } & \multicolumn{1}{c}{\cellcolor{customcolor}\textbf{15.0} } & \multicolumn{1}{c}{0.3 } & \multicolumn{1}{c}{14.1 } & \multicolumn{1}{c}{20.8 } & \multicolumn{1}{c}{\cellcolor{customcolor}6.8 } & \multicolumn{1}{c}{37.6 } & \multicolumn{1}{c}{86.8 } & \multicolumn{1}{c}{66.5 } & \multicolumn{1}{c}{55.3 } & \multicolumn{1}{c}{16.2 } & \multicolumn{1}{c}{40.6 } & \multicolumn{1}{c}{38.4 } & \multicolumn{1}{c|}{76.2 } & \multicolumn{1}{c}{\cellcolor{customcolor}7.1 } & \multicolumn{1}{c}{43.4 } & \multicolumn{1}{c}{35.7 } \\
    \multicolumn{1}{c|}{NOPS\cite{NOPS}} & \multicolumn{1}{c}{\cellcolor{customcolor}2.3 } & \multicolumn{1}{c}{\cellcolor{customcolor}27.8 } & \multicolumn{1}{c}{86.0 } & \multicolumn{1}{c}{89.9 } & \multicolumn{1}{c}{23.1 } & \multicolumn{1}{c}{24.5 } & \multicolumn{1}{c}{\cellcolor{customcolor}2.9 } & \multicolumn{1}{c}{3.1 } & \multicolumn{1}{c}{18.2 } & \multicolumn{1}{c}{30.1 } & \multicolumn{1}{c}{\cellcolor{customcolor}\textbf{16.3} } & \multicolumn{1}{c}{39.9 } & \multicolumn{1}{c}{90.7 } & \multicolumn{1}{c}{73.5 } & \multicolumn{1}{c}{61.0 } & \multicolumn{1}{c}{17.4 } & \multicolumn{1}{c}{49.8 } & \multicolumn{1}{c}{44.0 } & \multicolumn{1}{c|}{83.2 } & \multicolumn{1}{c}{\cellcolor{customcolor}12.4 } & \multicolumn{1}{c}{49.0 } & \multicolumn{1}{c}{41.2 } \\
    \multicolumn{1}{c|}{SNOPS \cite{SNOPS}} & \multicolumn{1}{c}{\cellcolor{customcolor}\textbf{4.7} } & \multicolumn{1}{c}{\cellcolor{customcolor}31.5 } & \multicolumn{1}{c}{84.6 } & \multicolumn{1}{c}{88.7 } & \multicolumn{1}{c}{22.8 } & \multicolumn{1}{c}{23.3 } & \multicolumn{1}{c}{\cellcolor{customcolor}8.2 } & \multicolumn{1}{c}{2.6 } & \multicolumn{1}{c}{17.9 } & \multicolumn{1}{c}{28.7 } & \multicolumn{1}{c}{\cellcolor{customcolor}15.1 } & \multicolumn{1}{c}{38.3} & \multicolumn{1}{c}{89.7 } & \multicolumn{1}{c}{72.5 } & \multicolumn{1}{c}{60.8 } & \multicolumn{1}{c}{16.1 } & \multicolumn{1}{c}{43.3 } & \multicolumn{1}{c}{45.7 } & \multicolumn{1}{c|}{82.9 } & \multicolumn{1}{c}{\cellcolor{customcolor}\textbf{14.9} } & \multicolumn{1}{c}{47.9 } & \multicolumn{1}{c}{40.9 } \\
    \multicolumn{1}{c|}{DASL\cite{DLAS}} & \multicolumn{1}{c}{\cellcolor{customcolor}2.6 } & \multicolumn{1}{c}{\cellcolor{customcolor}32.5 } & \multicolumn{1}{c}{88.7 } & \multicolumn{1}{c}{93.3 } & \multicolumn{1}{c}{28.1 } & \multicolumn{1}{c}{24.0 } & \multicolumn{1}{c}{\cellcolor{customcolor}0.1 } & \multicolumn{1}{c}{1.0 } & \multicolumn{1}{c}{23.7 } & \multicolumn{1}{c}{35.6 } & \multicolumn{1}{c}{\cellcolor{customcolor}15.3 } & \multicolumn{1}{c}{59.8 } & \multicolumn{1}{c}{93.2 } & \multicolumn{1}{c}{77.6 } & \multicolumn{1}{c}{61.4 } & \multicolumn{1}{c}{37.8 } & \multicolumn{1}{c}{56.6 } & \multicolumn{1}{c}{52.1 } & \multicolumn{1}{c|}{86.7 } & \multicolumn{1}{c}{\cellcolor{customcolor}12.6}  & \multicolumn{1}{c}{54.6 } & \multicolumn{1}{c}{45.8 } \\
        \multicolumn{1}{c|}{Ours} & \multicolumn{1}{c}{\cellcolor{customcolor}0.2 } & \multicolumn{1}{c}{\cellcolor{customcolor}\textbf{35.5} } & \multicolumn{1}{c}{89.1 } & \multicolumn{1}{c}{93.5 } & \multicolumn{1}{c}{28.7 } & \multicolumn{1}{c}{20.9 } & \multicolumn{1}{c}{\cellcolor{customcolor}2.1 } & \multicolumn{1}{c}{0.2 } & \multicolumn{1}{c}{21.9 } & \multicolumn{1}{c}{34.9 } & \multicolumn{1}{c}{\cellcolor{customcolor}9.5 } & \multicolumn{1}{c}{60.6 } & \multicolumn{1}{c}{93.0 } & \multicolumn{1}{c}{77.8 } & \multicolumn{1}{c}{63.2 } & \multicolumn{1}{c}{39.7 } & \multicolumn{1}{c}{63.7 } & \multicolumn{1}{c}{52.2 } & \multicolumn{1}{c|}{87.3 } & \multicolumn{1}{c}{\cellcolor{customcolor}11.8 } & \multicolumn{1}{c}{\textbf{55.0} } & \multicolumn{1}{c}{\textbf{45.9} } \\
      \bottomrule
       \bottomrule
    \end{tabular}
    
  }
  \vspace{-10pt}
  \label{tab2}
\end{table*}

\begin{figure*}[t]
  \centering
  \includegraphics[width=2.0\columnwidth]{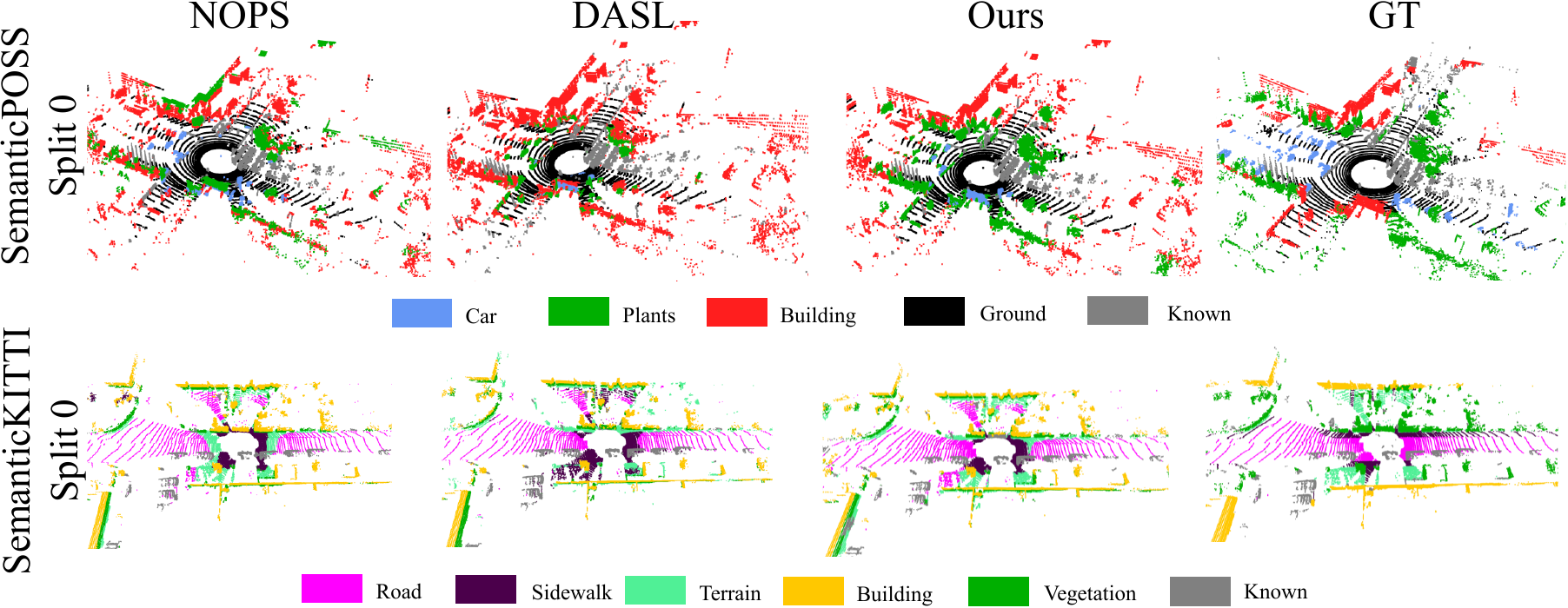}
  \vspace{-5pt}
  \caption{\textbf{Visualization comparison between our method, NOPS, and DASL on the SemanticPOSS and SemanticKITTI datasets.} As shown in the figure, our method achieves more accurate segmentation for novel classes, closely matching the ground truth. }
    \label{vis}
    \vspace{-15pt}
\end{figure*}

\textbf{SemanticKITTI Dataset.} The results in Table \ref{tab2} demonstrate our superior performance compared to previous methods on different splits of the SemanticKITTI Dataset. As shown in Table \ref{tab2}, our method achieves the highest average IoU across all four dataset splits, validating its effectiveness. This is particularly evident in the inference of certain novel classes. For example, on the pole category in Split 2, our method demonstrates a substantial improvement over previous approaches, supporting the rationale and effectiveness of using multiple base classes for novel class inference. Furthermore, on certain splits (e.g., Split 1 and Split 3), our results approach those of supervised methods, confirming the effectiveness of our approach for novel class discovery.

\textbf{Visualization Analysis.}  In Figure \ref{vis}, we present a visual comparison between NOPS, DASL, and our method. Our method shows significant improvements over the others. Specifically, as shown in the first row of Figure \ref{vis}, we can more accurately segment smaller objects in the point cloud, maintaining relatively complete shapes. For example, in the lower right of the first row, we accurately segment cars in the scene. This is achieved through our hypergraph structure built with Geometric-Aware Prototypes, which effectively reasons novel classes in the scene, resulting in precise segmentation outcomes. Additional segmentation visualization results are provided in the Supplementary Material.

\begin{figure*}[!t]
  \centering
  \includegraphics[width=2.0\columnwidth]{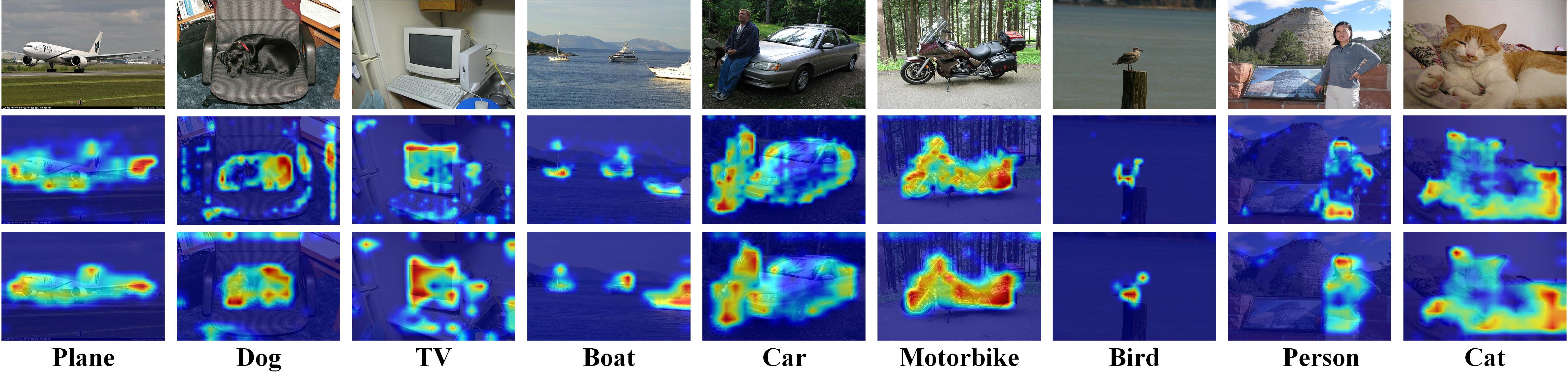}
\vspace{-10pt}
  \caption{Effectiveness of Geometric-Aware Prototype. The second row in the figure represents the prototype clustering method used in the EUMS \cite{EUMS}, while the third row illustrates geometric-aware prototypes used to build hypergraphs for novel class inference.}
  \vspace{-15pt}
    \label{heat}
\end{figure*}

\begin{table}[!t]
  \centering
   
  \caption{Ablation analysis of our model: GAP denotes Geometry-Aware Prototype, HSC is Hypergraph Structure Construction, and DHAM stands for Dynamic Hyperedge Adjustment Mechanism. }
  \vspace{-5pt}
  \resizebox{1.0\linewidth}{!}{
    \begin{tabular}{cccc|ccccc|c}
    \toprule
    \toprule
    \multicolumn{4}{c|}{Method} & \multicolumn{5}{c|}{Split0} & \multicolumn{1}{c}{Overall} \\
    \midrule
    \multicolumn{1}{c}{Baseline} & 
    \multicolumn{1}{c}{GAP} & 
    \multicolumn{1}{c}{HSC} & 
    \multicolumn{1}{c|}{DHAM} & 
    \multicolumn{1}{c}{Building} & 
    \multicolumn{1}{c}{Car} & 
    \multicolumn{1}{c}{Ground} & 
    \multicolumn{1}{c}{Plants} & 
    \multicolumn{1}{c|}{Avg} & 
    \multicolumn{1}{c}{Avg} \\
    \midrule
    $\checkmark$     &       &       &       & 50.8  & 3.2   & 35.9  & 48.7  & 34.7  & 23.2 \\
    $\checkmark$     & $\checkmark$     &       &       & 50.4  & 3.3   & 54.5  & 47.3  & 38.9  & 27.4 \\
    $\checkmark$     &       & $\checkmark$     &       & 55.7  & 8.9   & 68.4  & 40.8  & 43.5  & 32.8 \\
    $\checkmark$     & $\checkmark$     & $\checkmark$     &       & 56.7  & \textbf{6.5}   & 72.4  & \textbf{45.3}  & 45.2  & 34.2 \\
    $\checkmark$     & $\checkmark$     & $\checkmark$     & $\checkmark$     & \textbf{58.3} & 4.4 & \textbf{80.9} & \textbf{45.3} & \textbf{47.2} & \textbf{36.5} \\
    \bottomrule
    \bottomrule
    \end{tabular}%
    }
  \label{teb6}%
 \vspace{-20pt}
\end{table}%
 \begin{figure}[!h]
  \centering

  \includegraphics[width=1.0\columnwidth]{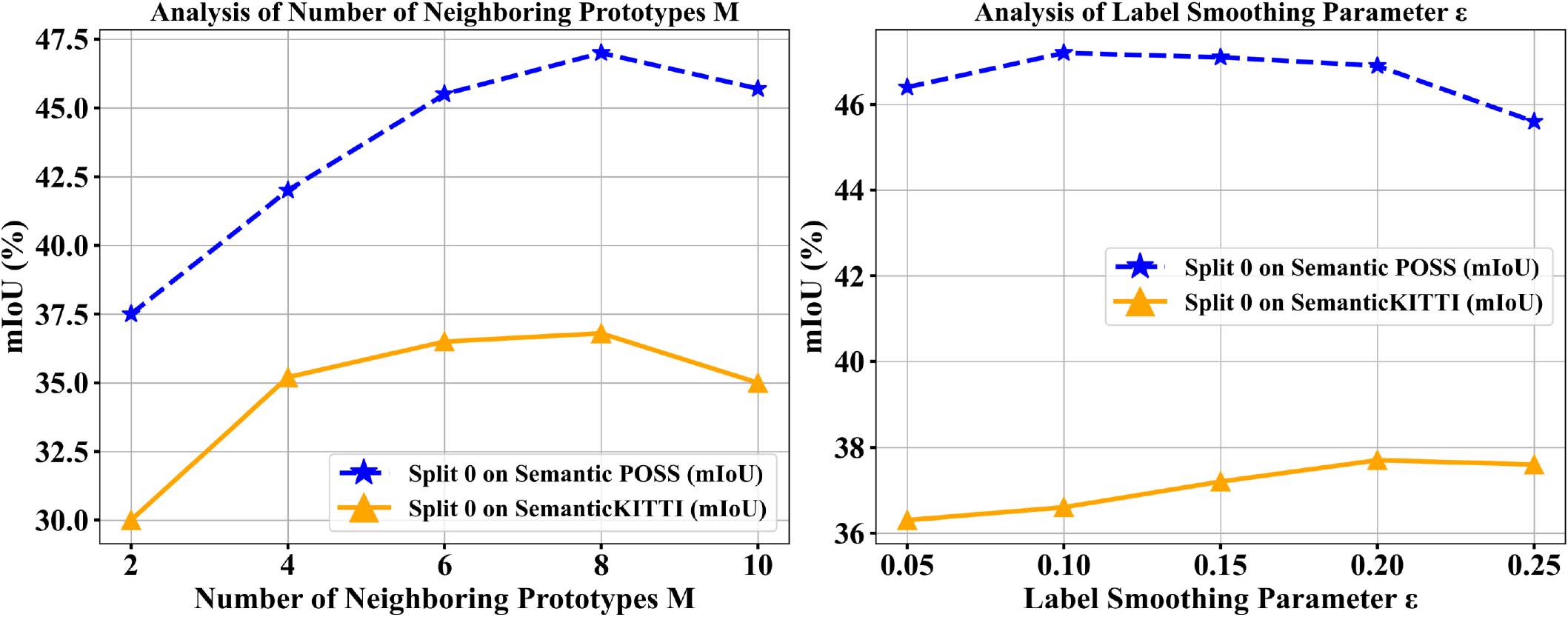}
\vspace{-15pt}
  \caption{Analysis of Neighboring Prototypes \( M \) in Hypergraph Construction. The label smoothing parameter \( \epsilon \) prevents overfitting by regularizing the dataset, enhancing the model's robustness.}
  \vspace{-15pt}
    \label{M}
\end{figure}
\subsection{Ablation Study}
\vspace{-5pt}
\textbf{Component Analysis.} We conduct ablation studies on the SemanticPOSS dataset to evaluate the effectiveness of three key modules in our framework: Geometric-Aware Prototypes, Hypergraph Structure Construction, and the Dynamic Hyperedge Adjustment Mechanism. Table \ref{teb6} reports results on the four novel classes in split 0, with ‘Avg’ denoting the mean mIoU across all four dataset splits.
The first row shows the baseline performance using a modified MinkowskiUNet-34C \cite{Mink}. Introducing Geometric-Aware Prototypes (second row) enhances spatial feature learning and improves novel class segmentation. The third row adds a hypergraph built with clustered prototypes, effectively modeling multi-relational interactions between base and novel classes. Replacing standard prototypes with geometric-aware ones (fourth row) further boosts performance, emphasizing the value of geometric information. Finally, the Dynamic Hyperedge Adjustment Mechanism (fifth row) refines the hypergraph across batches, enhancing prototype interactions and addressing class imbalance. These results validate the contribution of each component to the overall framework.


\textbf{Analysis of the Number of Neighboring Prototypes \( M \) in Hypergraph Construction.} In hypergraph construction, the number of neighboring prototypes \( M \) is a key hyperparameter. It controls the number of neighbors each prototype is connected to, preventing excessive interconnections and avoiding an overly complex or redundant hypergraph that could hinder subsequent inference.
As shown in Figure \ref{M}, on the relatively simple SemanticPOSS dataset, it achieves optimal performance when  \( M \) = 8. Notably, for the more complex SemanticKITTI dataset, which contains more categories and exhibits more intricate inter-class relationships, the model also attains its best results with the same hyperparameter setting. 
However, as \( M \) increases beyond an optimal point, excessive connections between prototypes can result in redundant semantic information, leading to performance degradation. This is evident in Figure \ref{M}, where both datasets show a decline in performance when \( M = 10 \).

\textbf{Analysis of the Label Smoothing Parameter \( \epsilon \).}  Given that SemanticPOSS and SemanticKITTI datasets exhibit class imbalance, label smoothing helps mitigate the imbalance between majority and minority classes, allowing the model to perform more evenly across different categories. As shown on the right of Figure \ref{M}, a larger smoothing coefficient reduces overfitting, especially on the class-rich SemanticKITTI dataset, improving inference; further details on \( \epsilon \) are in the Supplementary Material.

\textbf{Effectiveness of Geometric-Aware Prototype.} In our method, Geometric-Aware Prototypes are continuously updated during training, with one prototype per category. During inference, these prototypes, combined with the hypergraph, guide novel class prediction and point-wise classification in the point cloud. As shown in Figure \ref{heat}, the 3D visualization illustrates category-specific prototypes using distinct colors. However, due to the complexity of 3D scenes, their effectiveness is not fully apparent.
To better demonstrate their impact, we also visualize heatmaps on 2D projections. Comparing the second and third rows in Figure \ref{heat}, our geometric-aware prototypes capture object contours and local structures more effectively than traditional clustering, enhancing foreground focus and demonstrating improved precision and robustness for novel-class discovery.
We further validate Geometric-Aware Prototypes in the supplementary material by comparing performance using different prototypes (e.g., purely geometric or purely semantic).

\vspace{-5pt}
\section{Conclusion}
\vspace{-5pt}
For the task of Novel Class Discovery in Point Cloud Segmentation, we introduce a hypergraph structure to model the high-order interactions between classes, thereby facilitating cooperative reasoning for novel classes.  Additionally, we propose Geometric-Aware Prototypes to enhance the model's ability to capture spatial structural information. The hypergraph structure enables the modeling of multi-level interactions between novel classes and multiple base class prototypes, allowing these base prototypes to collaboratively infer a novel class prototype. This high-order interaction mechanism constructs a comprehensive semantic representation for novel classes, thereby enhancing their segmentation accuracy.
The significant performance improvements on the SemanticKITTI and SemanticPOSS datasets highlight the effectiveness of our method.

\noindent\textbf{Acknowledgment.}
This work is supported by the National Nature Science Foundation of China (Nos. 62376186, 62472333).

{
    \small
    \bibliographystyle{ieeenat_fullname}
    \bibliography{main}
}


\end{document}